\documentclass[twoside,11pt]{article}

\usepackage{jmlr2e}
\usepackage{comment}
\usepackage{amsmath}
\usepackage{float}
\usepackage{graphicx}

\graphicspath{ {./images/} }

\ShortHeadings{InterpretML - Draft}{Nori, Jenkins, Koch and Caruana}
\firstpageno{1}

\begin{document}

\title{InterpretML: A Unified Framework for Machine Learning Interpretability}

\author{\name Harsha Nori    \email hanori@microsoft.com    \\
        \name Samuel Jenkins \email sajenkin@microsoft.com  \\
        \name Paul Koch      \email paulkoch@microsoft.com  \\
        \name Rich Caruana    \email rcaruana@microsoft.com \\
       \addr Microsoft Corporation \\
       1 Microsoft Way \\
       Redmond, WA 98052, USA}

\editor{}
\maketitle


\begin{abstract}

InterpretML is an open-source Python package which exposes machine learning interpretability algorithms to practitioners and researchers. InterpretML exposes two types of interpretability -- \texttt{glassbox}, which are machine learning models designed for interpretability (ex: linear models, rule lists, generalized additive models), and \texttt{blackbox} explainability techniques for explaining existing systems (ex: Partial Dependence, LIME). The package enables practitioners to easily compare interpretability algorithms by exposing multiple methods under a unified API, and by having a built-in, extensible visualization platform. InterpretML also includes the first implementation of the Explainable Boosting Machine, a powerful, interpretable, glassbox model that can be as accurate as many blackbox models. The MIT licensed source code can be downloaded from \href{https://github.com/microsoft/interpret}{github.com/microsoft/interpret}. 

\end{abstract}


\section{Introduction}

As machine learning has matured into wide-spread adoption, building models that users can understand is becoming increasingly important. This can easily be observed in high-risk applications such as healthcare \citep{ahmad2018interpretable, caruana2015intelligible}, finance \citep{hajek2019interpretable, chen2018interpretable} and judicial environments \citep{Tan:2018:DAB:3278721.3278725, soundarajanequal}. Interpretability is also important in general applied machine learning problems such as model debugging, regulatory compliance, and human computer interaction.

We address these needs with InterpretML by exposing many state of the art interpretability algorithms under a unified API. This API covers two major interpretability forms: "glassbox" models, which are inherently intelligible and explainable to the user, and "blackbox" interpretability, methods that generate explanations for any machine learning pipeline, no matter how opaque it is. This is further supported with interactive visualizations and a built-in dashboard designed for interpretability algorithm comparison. InterpretML is MIT licensed, and emphasizes extensibility and compatibility with popular open-source projects such as scikit-learn \citep{scikit-learn} and Jupyter Notebook environments \citep{Kluyver:2016aa}.

\section{Package Design}

InterpretML follows four key design principles that influence its architecture and API.

\textit{Ease of comparison.} Make it as easy as possible to compare multiple algorithms. ML interpretability is in its infancy, and many algorithmic approaches have emerged from research, each of which has pros and cons. Comparison is critical to find the algorithm that best suits the users' needs. InterpretML enables this by enforcing a scikit-learn style uniform API, and providing a visualization platform centered around algorithmic comparison. 

\textit{Stay true to the source.} Use reference algorithms and visualizations as much as possible. Our goal is to expose interpretability algorithms to the world, in their most accurate form. 

\textit{Play nice with others.} Leverage the open-source ecosystem, and don't reinvent the wheel. InterpretML is highly compatible with popular projects like Jupyter Notebook and scikit-learn, and builds off of many libraries like plotly, lime, shap, and SALib.

\textit{Take what you want.} Use and extend any component of InterpretML without pulling in the whole framework. For example, it's possible to produce a computationally intensive explanation on a server, without InterpretML's visualization and its related dependencies.

\begin{figure}[H]
    \centering
    \includegraphics[scale=0.6]{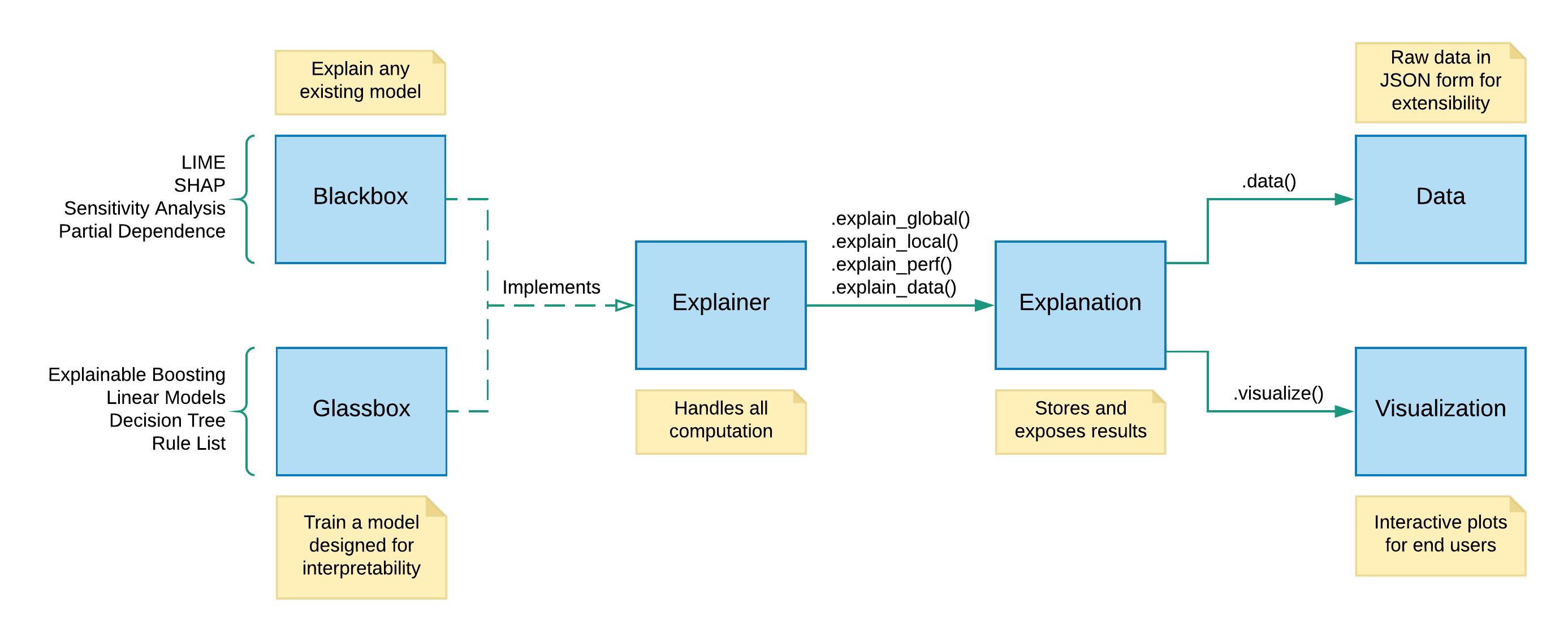}
    \includegraphics[scale=0.5]{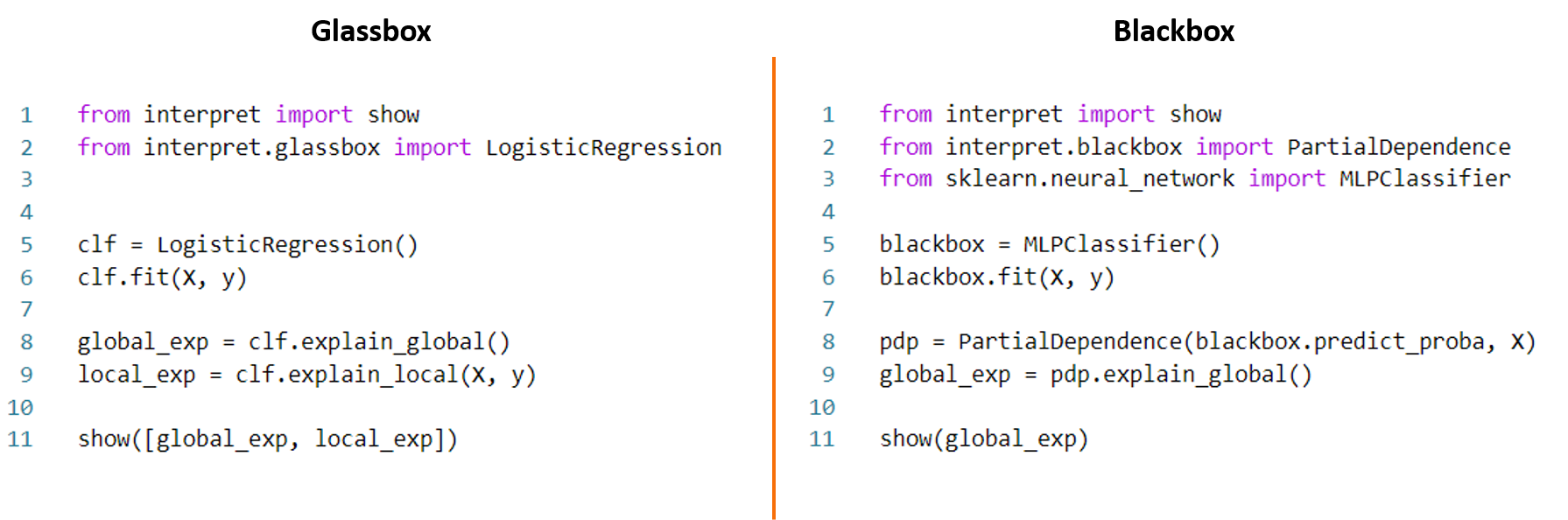}
    \caption{API architecture and code examples}
    \label{fig:interpret_arch}
\end{figure}

The code architecture and unified API is best expressed in Figure \ref{fig:interpret_arch}, providing an overview and relevant example code.

\section{Explainable Boosting Machine}

As part of the framework, InterpretML also includes a new interpretability algorithm -- the Explainable Boosting Machine (EBM). EBM is a glassbox model, designed to have accuracy comparable to state-of-the-art machine learning methods like Random Forest and Boosted Trees, while being highly intelligibile and explainable. EBM is a generalized additive model (GAM) of the form:
\setlength{\abovedisplayskip}{0pt}%
\setlength{\belowdisplayskip}{0pt}%
\setlength{\abovedisplayshortskip}{0pt}%
\setlength{\belowdisplayshortskip}{0pt}%
\begin{align*}
g(E[y]) = \beta_0 + \sum f_j(x_j)
\end{align*}
\noindent
where \textit{g} is the link function that adapts the GAM to different settings such as regression or classification. EBM has a few major improvements over traditional GAMs \citep{hastie1987generalized}. First, EBM learns each feature function  f\textsubscript{j} using modern machine learning techniques such as bagging and gradient boosting. The boosting procedure is carefully restricted to train on one feature at a time in round-robin fashion using a very low learning rate so that feature order does not matter.  It round-robin cycles through features to mitigate the effects of co-linearity and to learn the best feature function f\textsubscript{j} for each feature to show how each feature contributes to the model's prediction for the problem. Second, EBM can automatically detect and include pairwise interaction terms of the form: 
\setlength{\abovedisplayskip}{0pt}%
\setlength{\belowdisplayskip}{0pt}%
\setlength{\abovedisplayshortskip}{0pt}%
\setlength{\belowdisplayshortskip}{0pt}%
\begin{align*}
g(E[y]) = \beta_0 + \sum f_j(x_j) + \sum f_{i_j}(x_i,x_j)
\end{align*}
which further increases accuracy while maintaining intelligibility. 
EBM is a fast implementation of the GA\textsuperscript{2}M algorithm \citep{LouCGH13}, written in C++ and Python. The implementation is parallelizable, and takes advantage of \texttt{joblib} to provide multi-core and multi-machine parallelization. The algorithmic details for the training procedure, selection of pairwise interaction terms, and case studies can be found in \citep{LouCG12, LouCGH13, caruana2015intelligible}.

EBMs are highly intelligible, because the contribution of each feature to a final prediction can be visualized and understood by plotting f\textsubscript{j}. Because EBM is an additive model, each feature contributes to predictions in a modular way that makes it easy to reason about the contribution of each feature to the prediction. 

\begin{figure}[ht]
    \centering
    \includegraphics[scale=0.25]{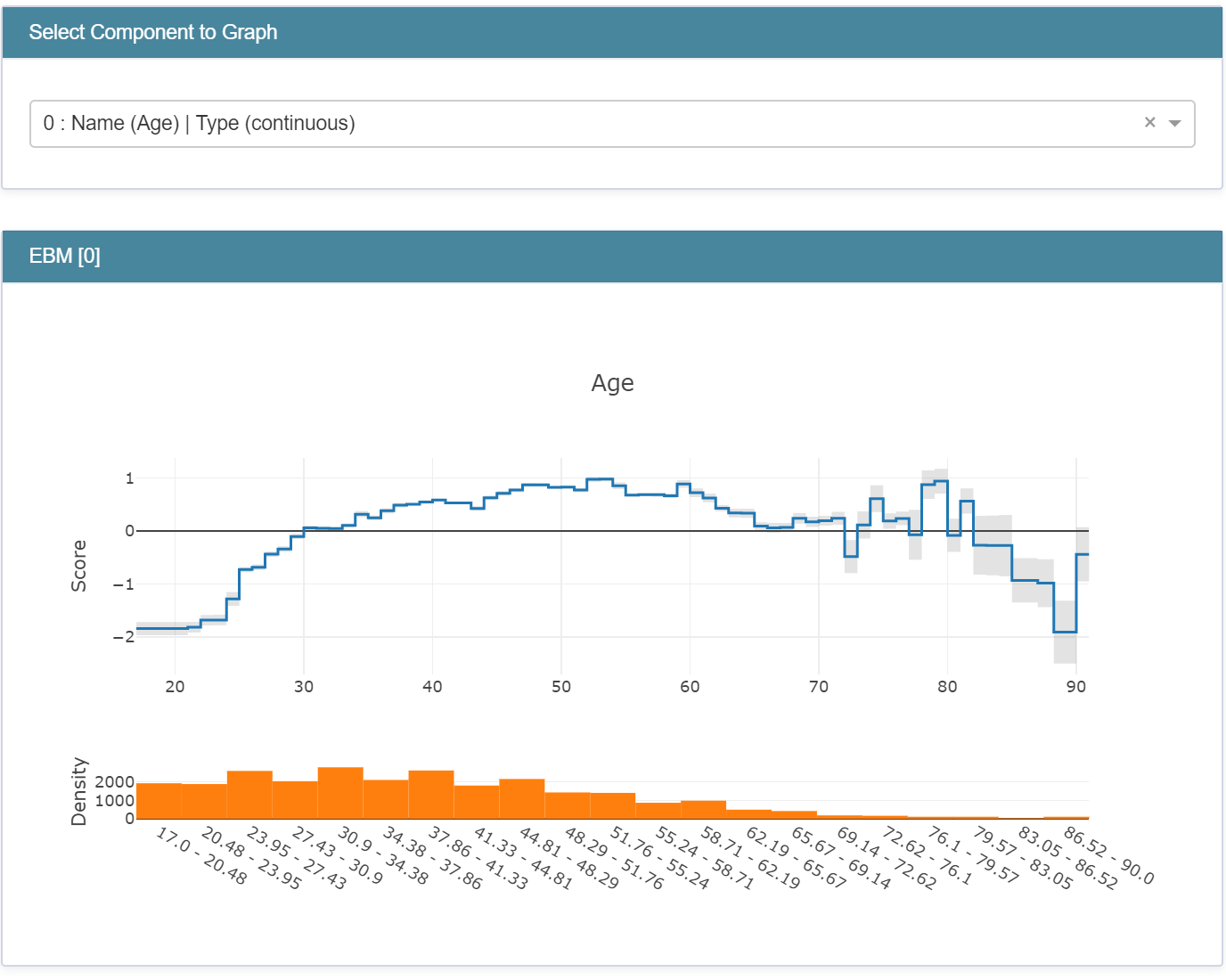}
    \includegraphics[scale=0.25]{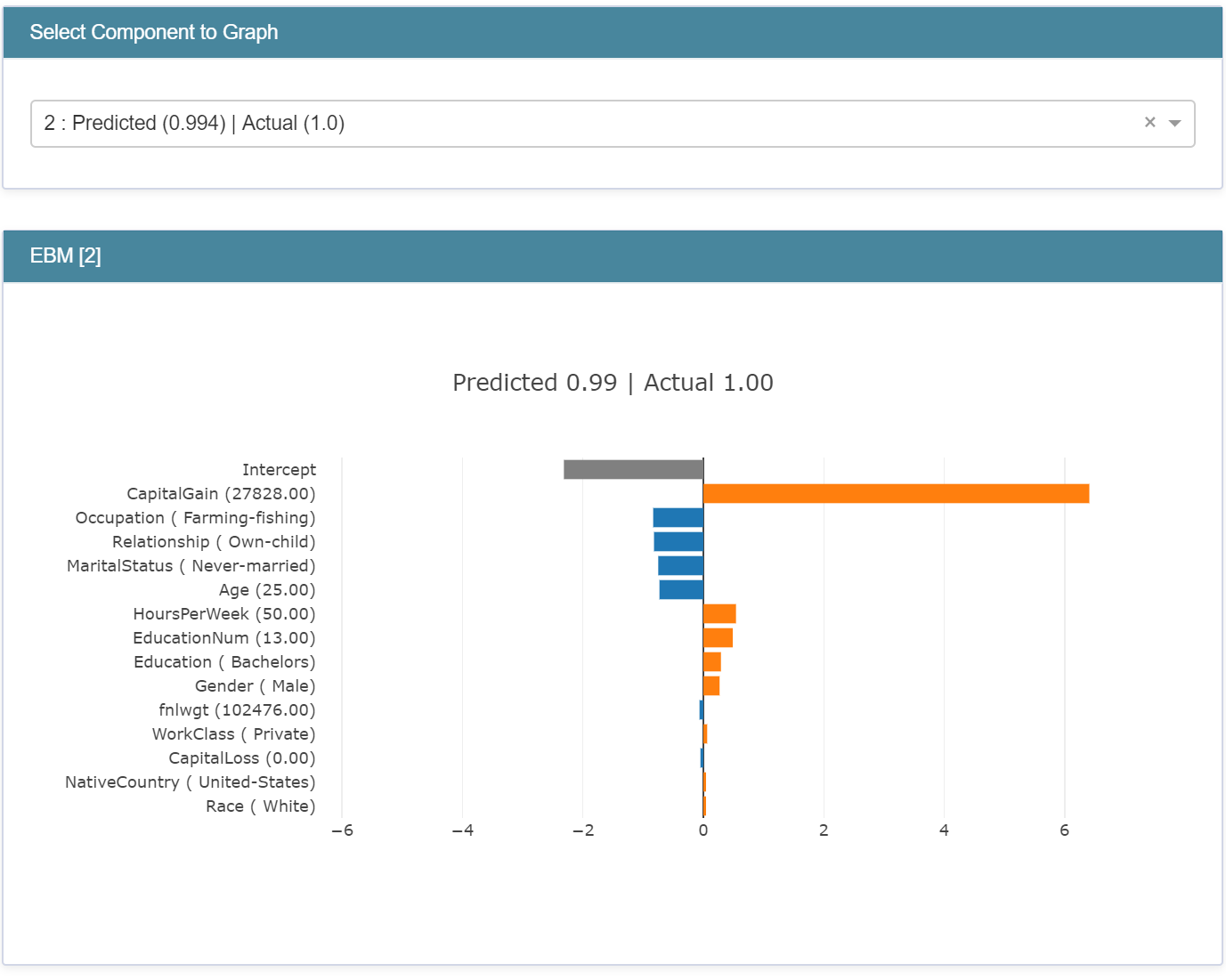}
    \caption{Left: Function f\textsubscript{Age}. As 'Age' increases from 20 to 50, p(1) increases significantly.
    Right: Individual prediction, where the 'CapitalGain' feature dominates.}
    \label{fig:ebmglobal}
\end{figure}

To make individual predictions, each function f\textsubscript{j} acts as a lookup table per feature, and returns a term contribution. These term contributions are simply added up, and passed through the link function \textit{g} to compute the final prediction. Because of the modularity (additivity), term contributions can be sorted and visualized to show which features had the most impact on any individual prediction. 

\begin{figure}[ht]
\begin{tabular}{ |p{3.2cm}||p{1.5cm}|p{1.5cm}|p{2cm}|p{2cm}|p{2.1cm} | }
 \hline
 \multicolumn{6}{|c|}{Classification Performance (AUROC)} \\
 \hline
 Model & heart-disease (303, 13) & breast-cancer (569, 30) & telecom-churn (7043, 19) & adult-income (32561, 14)  & credit-fraud (284807, 30)\\
 \hline
 EBM & \textbf{0.916} & \textbf{0.995} & \textbf{0.851} & \textbf{0.928} & 0.975\\
 LightGBM & 0.864 & 0.992 & 0.835 & \textbf{0.928} & 0.685\\
 Logistic Regression & 0.895 & \textbf{0.995} & 0.804 & 0.907 & 0.979\\
 Random Forest & 0.89 & 0.992 & 0.824 & 0.903 & 0.95 \\
 XGBoost & 0.87 & \textbf{0.995} & 0.85 & 0.922 & \textbf{0.981}\\
 \hline
\end{tabular}
\caption{Classification performance for models across datasets (rows, columns).}
\end{figure}

\begin{figure}[ht]
    \centering
    \includegraphics[scale=0.28]{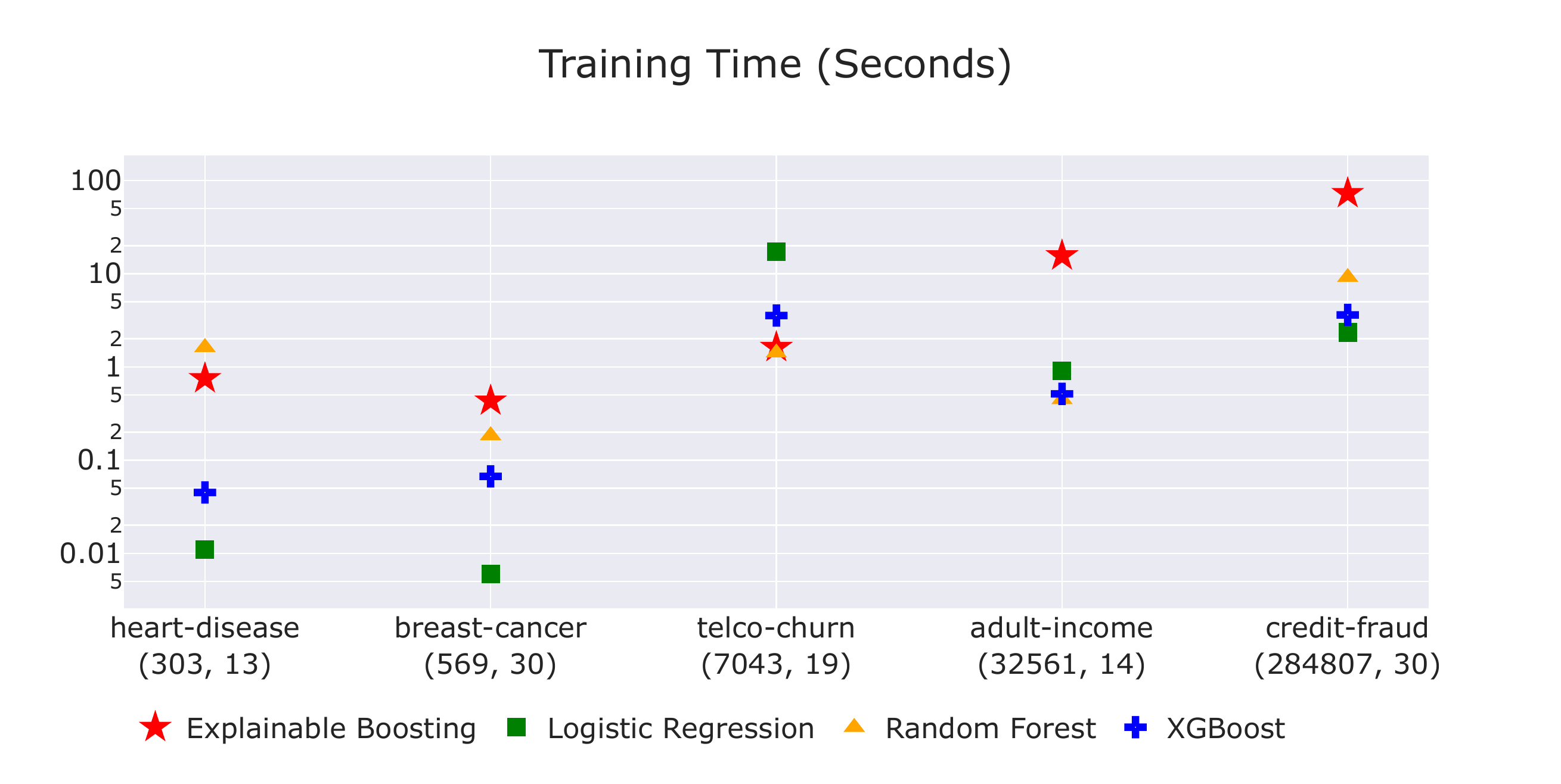}
    \includegraphics[scale=0.28]{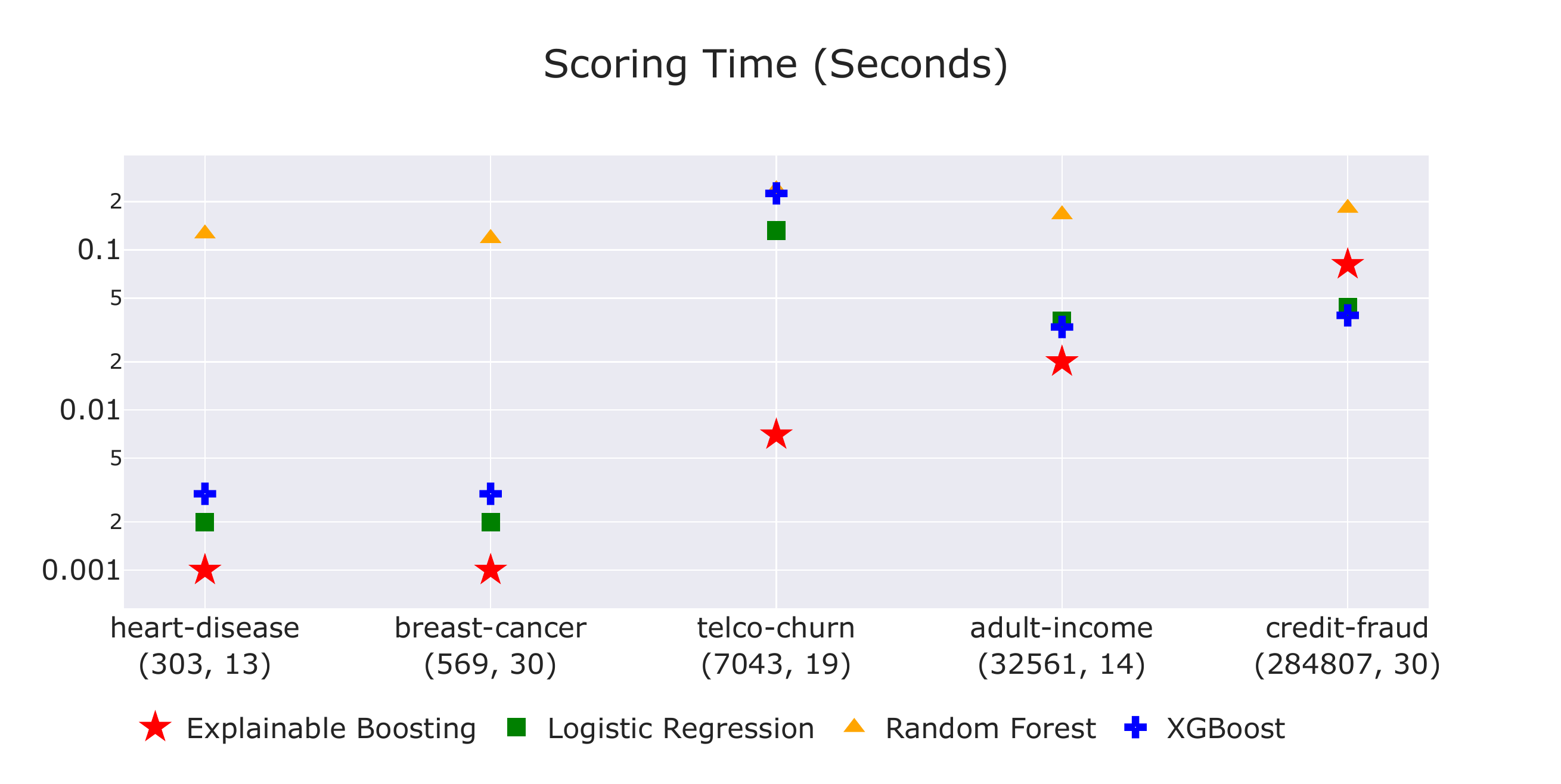}
    \caption{Computational performance for models across datasets (rows, columns).}
    \label{fig:comp_perf}
\end{figure}

In terms of predictive power, EBM often performs surprisingly well, and is comparable with state of the art methods like Random Forest and XGBoost.\footnote{All models were trained with their default parameters. EBM's current default parameters are chosen for computational speed, to enable ease of experimentation. For the best accuracy and interpretability, we recommend using reference parameters: 100 inner bags, 100 outer bags, 5000 epochs, and a learning rate of 0.01.} To keep the individual terms additive, EBM pays an additional training cost, making it somewhat slower than similar methods. However, because making predictions involves simple additions and lookups inside of the feature functions f\textsubscript{j}, EBMs are one of the fastest models to execute at prediction time. EBM's light memory usage and fast predict times makes it particularly attractive for model deployment in production.  

\acks{We would like to acknowledge everyone on our \href{https://github.com/microsoft/interpret/blob/master/ACKNOWLEDGEMENTS.md}{acknowledgements.md} file for their support on this project. } We also depend on many amazing software packages and research: scikit-learn \citep{scikit-learn}, plotly \citep{plotly}, lime \citep{lime}, shap \citep{NIPS2017_7062}, SALib \citep{Herman2017}, partial dependence \citep{friedman2001}, Jupyter \citep{Kluyver:2016aa}, pandas \citep{mckinneypandas}, and more.


\newpage
\newpage

\nocite{*}
\bibliography{sample}

\end{document}